\documentclass{article}

\PassOptionsToPackage{numbers, compress}{natbib}

\usepackage[final]{neurips_2024}

\usepackage[utf8]{inputenc} 
\usepackage[T1]{fontenc}    
\usepackage{hyperref}       
\usepackage{url}            
\usepackage{booktabs}       
\usepackage{amsfonts}       
\usepackage{nicefrac}       
\usepackage{microtype}      
\usepackage{xcolor}         

\usepackage{times}
\usepackage{latexsym}
\usepackage{multirow}
\usepackage{tikz}
\usepackage{graphicx}
\usepackage{epigraph}
\usepackage{mathtools}
\usepackage{changepage}
\usepackage{listings}
\usepackage{amssymb}

\title{Can Generic LLMs Help Analyze Child-Adult Interactions Involving Children with Autism in Clinical Observation?}

\author{%
  Tiantian Feng$^{1}$, Anfeng Xu$^{1}$, Rimita Lahiri$^{1}$, 
  \\ \textbf{Sudarsana Reddy Kadiri$^{1}$, Helen Tager-Flusberg$^{2}$, So Hyun Kim$^{3}$,}
  \\\textbf{Somer Bishop$^{4}$, Catherine Lord$^{5}$, Shrikanth Narayanan$^{4}$}\\ 
$^1$University of Southern California, $^2$Boston University, $^3$Korea University, 
\\$^4$University of California, San Francisco, $^5$ University of California, Los Angeles\\
{\small\texttt{tiantiaf@usc.edu}}\\}

\begin{document}

\maketitle

\begin{abstract}
  Large Language Models (LLMs) have shown significant potential in understanding human communication and interaction. However, their performance in the domain of child-inclusive interactions, including in clinical settings, remains less explored. In this work, we evaluate generic LLMs' ability to analyze child-adult dyadic interactions in a clinically relevant context involving children with ASD. Specifically, we explore LLMs in performing four tasks: classifying child-adult utterances, predicting engaged activities, recognizing language skills and understanding traits that are clinically relevant. Our evaluation shows that generic LLMs are highly capable of analyzing long and complex conversations in clinical observation sessions, often surpassing the performance of non-expert human evaluators. The results show their potential to segment interactions of interest, assist in language skills evaluation, identify engaged activities, and offer clinical-relevant context for assessments.
\end{abstract}

\section{Introduction}
\label{sec:introduction}

Humans develop unique language abilities at different stages of life, each associated with significant milestones in communication skills. For example, children acquire basic language skills and vocabulary in early childhood. As they progress into adolescence, they learn to include more complex sentence structures and enriched vocabulary. However, language development may differ from typical patterns for individuals with Autism Spectrum Disorder (ASD) \cite{lord2020autism,volden1991neologisms}. ASD is a neurodevelopmental disorder that impairs social interactions and delays the acquisition and use of language.  Understanding behaviors associated with ASD benefits clinicians in providing effective interventions to support their needs \cite{bone2015applying}. One standard protocol used for this purpose is the Autism Diagnostic Observation Schedule (ADOS) \cite{lord2000autism}. It involves semi-structured dyadic interactions between a child and a clinician to evaluate their abilities and behaviors in communication. Understanding child-inclusive interactions in these clinical assessments creates unique opportunities for ASD diagnosis.


\begin{figure}[ht]
  \centering
  \includegraphics[width=\linewidth]{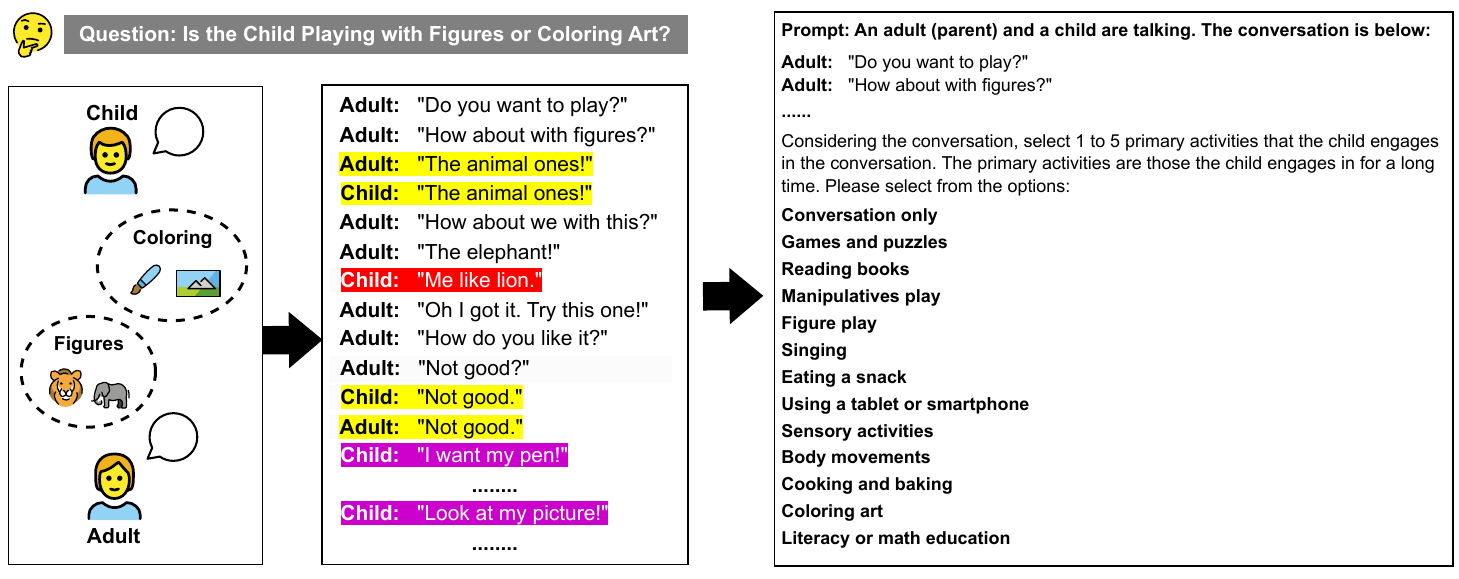}
  \caption{An illustration of a part of the child-adult conversation \textit{with the protocols in Remote-NLS}. All examples in this paper are scripted instead of raw due to IRB restrictions. Blue, red, and purple denote repetitions, language skills, and engaged activity, respectively. The figures on the right show a prompt template for selecting the activities that are engaged from the transcripts.}
  \label{fig:introduction}
  \vspace{-2.5mm}
\end{figure}

This work presents one of the earliest efforts to evaluate LLM's \cite{zhao2023survey} ability to understand child-adult dyadic conversations in clinical settings. Specifically, we introduce 4 tasks that are relevant to the automatic understanding of the conversations. These tasks include both utterance-level and session-level understanding: 1) Recognizing the speaker of the utterance (child/adult); 2) Recognizing engaged activities/observations from the entire conversation; 3) Predicting the child's traits (e.g., age); 4) Predicting the child's language skills. 

Our tasks are challenging and important in understanding child-adult dyadic interactions in clinical settings, involving processing long conversations presented in Fig~\ref{fig:introduction}. Specifically, these interactions often include repetitions, pauses, topic shifts, unexpected child responses, and varying language skills for children with ASD. Fig~\ref{fig:introduction} shows an activity recognition example with topic switching and constant repetitions between the adult and the child, from the adult suggesting figure play to the child choosing coloring art. We also demonstrate a template (on the right) to prompt the LLMs to select engaged activities throughout the session. We would highlight that this interaction type, involving children with ASD, is largely unexplored in LLMs, creating a low-resource context for evaluating LLMs.

In summary, our evaluation highlights the promise of LLMs to analyze child-adult interactions (in transcripts) involving children with autism. First, LLMs are highly capable of understanding the speaker of the utterances, showing their potential to consider the context of the conversation and recognize the relevant contents of the child. Moreover, LLMs show promise in predicting activities/observations in a session and children's language skills, even outperforming non-experts in performing these tasks. This implies their ability to assist clinicians in segmenting the interactions of interest and evaluating the child's language skills. However, we observe that LLMs frequently produce hallucinated outputs.

\section{Datasets}
\label{sec:dataset}

We consider two datasets in our experiments. These two datasets are the Remote-NLS \cite{butler2022remote} and ADOSMod3 datasets, which represent naturalistic and semi-structured assessments, respectively. Both datasets involve child-adult dyadic interactions collected under ASD-relevant contexts. We use the transcripts annotated by human experts to ensure the transcription quality. We comply with the data usage in accordance with the relevant IRB and DUAs from the original data owners. The data statistics, including demographics of children and utterance numbers, are in Table~\ref{tab:dataset}.

\vspace{0.7mm}
\noindent\textbf{Remote-NLS} \cite{butler2022remote} includes 89 Zoom videos of 15-minute child-adult interactions at home. Parents were instructed to choose a few activities (e.g., games) to elicit the child's Natural Language Samples (NLS) \cite{barokova2020commentary} for clinical assessments. The majority of the children are minimally verbal and diagnosed with ASD. Experts transcribed the videos and categorized children's language skills into pre-verbal, first words, and word combinations based on \cite{tager2009defining}. Specifically, \textbf{pre-verbal} is a stage where children rely on vocal (babble) and gestural means to communicate. Children in this stage are typically around 6-12 months. Moreover, \textbf{the first words} represent the phase where children are able to apply a single word to refer to objects and events. Finally, in \textbf{word combinations}, children develop their vocabulary and combine words to communicate about objects. The complete descriptions of the language skills are demonstrated in the prompt template shown in Figure~\ref{fig:language_prompt}. This dataset evaluates the spontaneous spoken language of children with ASD in a naturalistic context.

\vspace{0.7mm}
\noindent \textbf{ADOSMod3} dataset involves child-clinician dyadic interactions following the ADOS-2 protocol \cite{lord2000autism, bishop2017autism}. Specifically, we focus on the ``Emotion" and ``Social Difficulties and Annoyance" subtasks, with recordings from the children participating in both subtasks, each of which lasts around 3 minutes. The dataset includes recordings from 164 children following the administration of module 3 designed for verbally fluent children. The participants are verbally fluent children with and without ASD symptoms. Children without ASD may have other types of diagnosis, including ADHD and symptoms/mood/intellectual disorder. In this dataset, we evaluate the language skill using the standard E1 scale of ADOS evaluation, and the definitions of the language skills are demonstrated in Figure~\ref{fig:language_prompt}.

\section{Task Designs}

After consulting and discussing with experts in the field, we designed four tasks to benchmark LLMs' capabilities in understanding child-adult interactions involving children with ASD: classifying child-adult utterances, recognizing engaged activities, and predicting child traits and language skills. We would highlight that the child-adult classification is at the utterance level, while the remaining tasks are at the session level, where the entire transcript of the conversation is included in the prompt message. Our aim is that LLMs must first understand the nuances involved in lengthy conversations so that they can provide useful and meaningful analysis to clinicians.

\vspace{0.7mm}
\noindent \textbf{Child-adult Utterance Classification.} 
This task classifies utterances as child or adult, evaluating the potential for automatic speaker segmentation in transcripts without reliable speech diarization, which benefits the child's behavior assessment by highlighting their spoken contents. For example, we would highlight that our current prompt experiment can be extended to include LLMs to correct diarization results from ASR systems \cite{wang2024diarizationlm}.
In Table~\ref{table:utterance}, we present an example to classify the utterance "The elephant" as either child or adult based on the conversation context, highlighting challenges in understanding these unstructured conversations.

\begin{figure}[t]
  \centering
  \includegraphics[width=\linewidth]{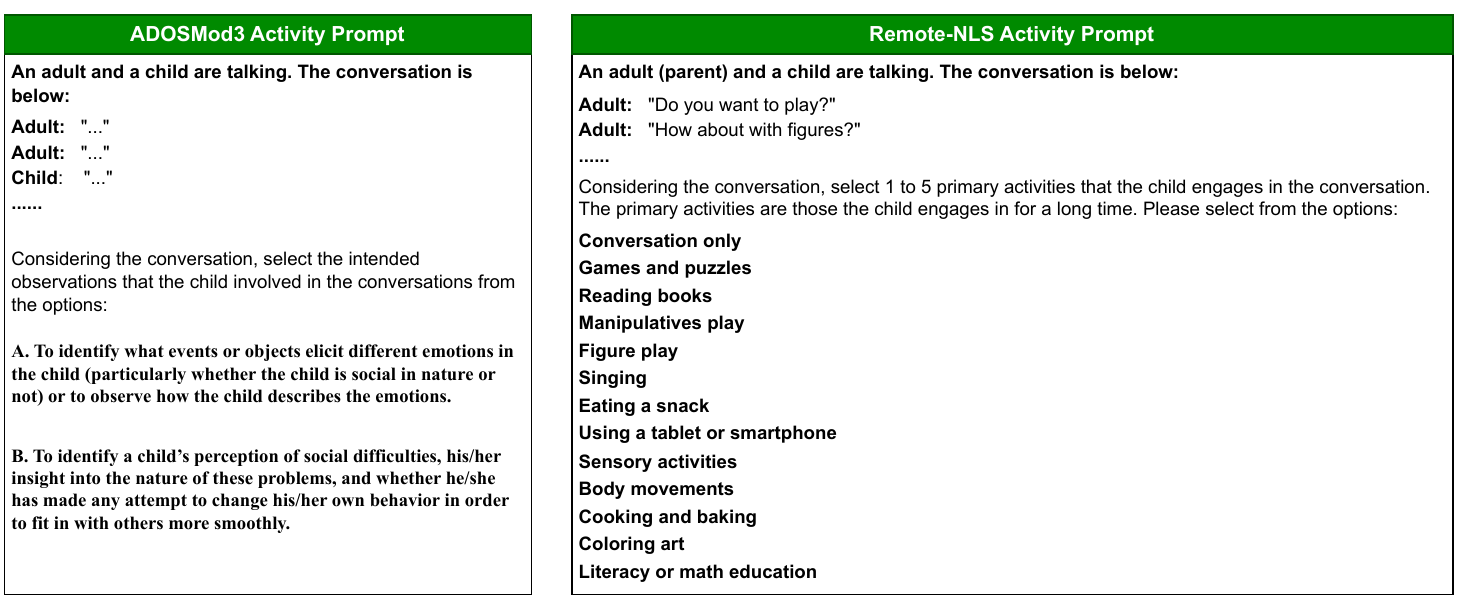}
  \caption{The figure shows a prompt template for recognizing activities in both Remote-NLS and ADOSMod3 datasets.}
  \label{fig:activity_prompt}
\end{figure}

\vspace{0.7mm}
\noindent \textbf{Activities/Observation Task Recognition.} 
The evaluation involves selecting engaged activities in child-adult dyadic interactions, with multiple correct answers from 13 options in the Remote-NLS dataset as demonstrated in Figure~\ref{fig:activity_prompt}. For the ADOSMod3 dataset, the task is to predict whether the conversation involves "Emotion" or "Social Difficulties and Annoyance" observations. This task helps determine if LLMs can segment conversations by specific activities/observations for clinical evaluations. The detailed prompt template is shown in Figure~\ref{fig:activity_prompt}.

\begin{table}[t]
  \centering
  \caption{Remote-NLS and ADOSMod3 dataset statistics. }
  \footnotesize
  \resizebox{0.98\textwidth}{!}{
  \begin{tabular}{lclc}
    \toprule

    \multicolumn{2}{c}{\textbf{Remote-NLS}} & \multicolumn{2}{c}{\textbf{ADOSMod3}}  \\
    
    \multicolumn{1}{c}{\textbf{Category}} & \textbf{Statistics} & \multicolumn{1}{c}{\textbf{Category}} & \textbf{Statistics}  \\
    \cmidrule(lr){1-2} \cmidrule(lr){3-4}
    Age (month) & Range: $49-95$ , $\mu\pm\sigma$: $75\pm13$ & Age (month) & Range: $43-158$ , $\mu\pm\sigma$: $104 \pm 29$ \\
    Gender &  $70$ males, $19$ females & Gender &  $121$ males, $43$ females \\
    Utt. Per Session & Adult: $258\pm82$; Child: $104\pm64$ & Utt. Per Session & Adult: $31 \pm 17$; Child: $27 \pm 21$ \\
    \bottomrule
  \end{tabular}
  }
\label{tab:dataset}
\end{table}

\begin{figure}[t]
  \centering
  \includegraphics[width=\linewidth]{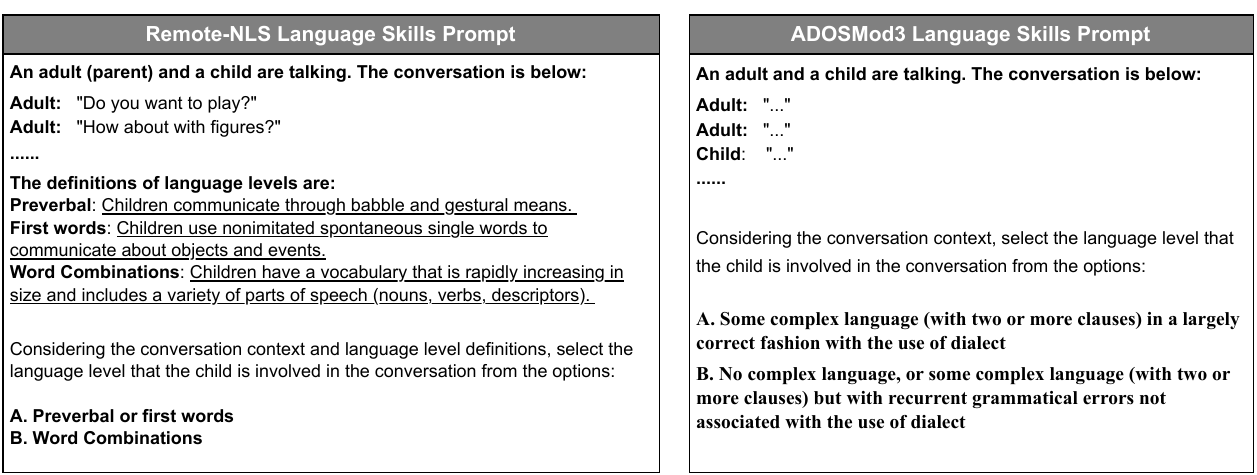}
  \caption{The figure shows a prompt template for recognizing language skills in both Remote-NLS and ADOSMod3 datasets.}
  \label{fig:language_prompt}
\end{figure}

\vspace{0.7mm}
\noindent \textbf{Language Skill.} 
Two unique codes evaluate language skills in the Remote-NLS and ADOSMod3 datasets. For Remote-NLS, the options are "Preverbal or first words" and "Word Combinations" following \cite{tager2009defining}. The ADOSMod3 code is described in Figure~\ref{fig:language_prompt}, and we evaluate with children from 4-10 years. This task tests if LLMs can assist clinicians in evaluating language skills, which is a relevant ASD indicator.

\begin{figure}[t]
  \centering
  \includegraphics[width=\linewidth]{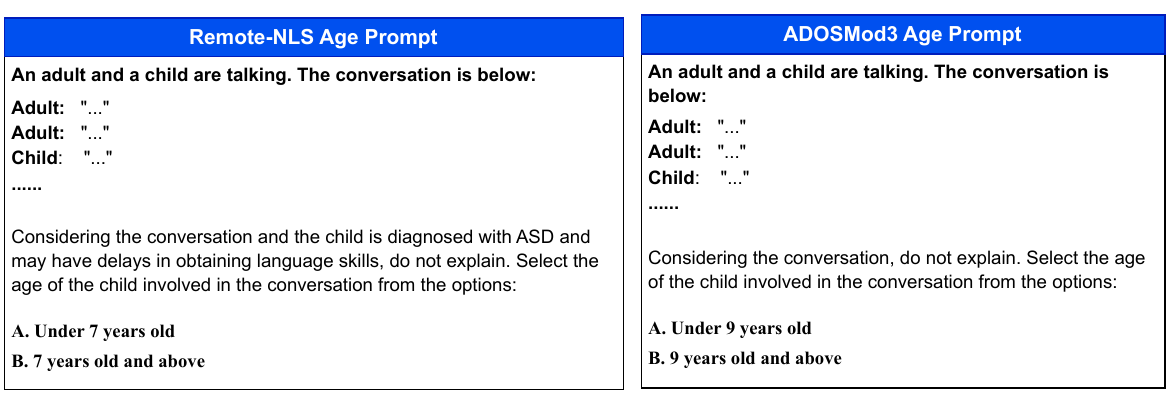}
  \caption{The figure shows a prompt template for classifying age ranges in both Remote-NLS and ADOSMod3 datasets.}
  \label{fig:age_prompt}
\end{figure}

\vspace{0.7mm}
\noindent \textbf{Traits Prediction.} 
The task is to predict the binarized age range (e.g., "Under 9 years old" or "9 years old and above" in the ADOSMod3). Since children with ASD may experience delayed language acquisition, knowing the child's age helps clinicians understand their conditions. The detailed prompt template for age range classification is shown in Figure~\ref{fig:age_prompt}.

\section{Experiments}
\label{sec:experiments}

\subsection{Evaluation Methods}

\vspace{0.7mm}
\noindent \textbf{LLMs.} \textit{Due to strict IRB considerations in using these data in ASD-relevant context, we avoid using Cloud-based LLMs such as GPT4o \footnote{https://openai.com/index/hello-gpt-4o/}}. Instead, we use locally deployable open-source LLMs. We use the model weights downloaded from the Huggingface for all the models used in this work. Moreover, we load the weights with float16 in all experiments. Specifically, we use Mistral-7B V0.2 Instruction \cite{jiang2023mistral}, LLaMa 2-7B Chat \cite{touvron2023llama}, LLaMa 2-13B Chat \cite{touvron2023llama}, LLaMa 3-8B Instruct \footnote{https://llama.meta.com/llama3/}, Qwen1.5-7B Chat \cite{bai2023qwen}, and Qwen1.5-14B Chat \cite{bai2023qwen} models for the experiments. Our selection criteria included models with parameter sizes below 15B to ensure compatibility with a wide range of computing facilities. We ran the models with a temperature of 0.01.

\vspace{0.7mm}
\noindent \textbf{Non-expert Human Evaluation.} Following the work in~\cite{hessel2023androids}, we include human evaluations to compare with LLMs. Specifically, we invite co-authors who have proper IRB training and are unfamiliar with child-adult interactions in clinical contexts to evaluate. We choose not to include crowd-workers due to IRB restrictions. Each evaluator reviews 50 utterance classification examples (10-shot). Moreover, they review ten randomly selected transcriptions from the Remote-NLS to predict engaged activities and language skills. We note that non-expert evaluations, such as on language skills, \textit{represent the general population’s ability to perform these tasks but not clinicians.}

\vspace{0.7mm}
\noindent \textbf{Trained Baseline.} We include baseline in utterance classification tasks fine-tuned with RoBERTa \cite{liu2019roberta}. The training baselines for classifying child-adult classification involve using RoBERTa \cite{liu2019roberta} to extract CLS embeddings from utterances. We train the classification using cross-validation with a 3-fold cross-fold evaluation. We apply 2-layer MLP layers as the classifier. Each experiment adopts a learning rate from ${0.0001, 0.00005}$ and a maximum training epoch of 20. The average F1 score is reported by averaging the F1 score across different training folds. As for the session-level understanding tasks, it is not feasible to train BERT-life models with extensive input text length, so we decided to compare the performance of LLMs with human evaluations. 

\subsection{Computing Infrastructure}
We conducted all the experiments on a server with 4 A6000 GPUs, a 128-core Processor, and 256G memory. The GPU hours to run each session-level understanding task for a single LLM are typically within one hour, while the GPU hours to run utterance-level understanding tasks for a single LLM are approximately under 4 hours.

\subsection{Experiment Details and Metrics}

In classifying child-adult utterances, we sample every fifth utterance from each transcript for evaluation due to the large number of total utterances. We use both 0-shot and context-based prompts, with the latter including adult-child utterance examples, which naturally is a few-shot prompt. We also exclude empty utterances. Moreover, we evaluate session-level understanding tasks using the entire session transcript with speaker labels. We report LLM performance using the F1 score. 


\newcommand{\zshotres}{\hspace*{.4in}\rotatebox[origin=c]{180}{$\Lsh$}\xspace}

\begin{table*}[t]
    \caption{Results for child-adult classification with the single utterance. \textbf{Bold} indicate the best results in each dataset with LLMs. 5-shot includes conversation contexts of 5 history utterances with child and adult labels.}
    \footnotesize
    \centering
    \begin{minipage}{0.48\textwidth}
        \centering
        \includegraphics[width=\linewidth]{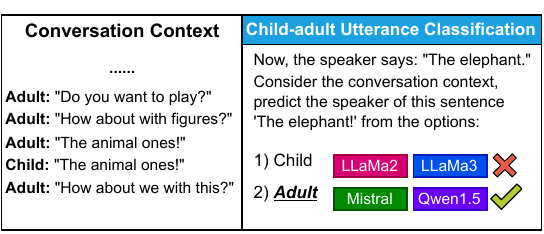}
    \end{minipage}
    \begin{minipage}[t]{0.51\textwidth}
        \centering
        \resizebox{0.85\textwidth}{!}{
            \begin{tabular*}{\linewidth}{p{2.65cm}p{1.8cm}p{1.8cm}}

                \footnotesize
                
                & \textbf{ADOSMod3} & \textbf{Remote-NLS} \\
                \midrule
                
                LLaMa2-7B(0-shot) & 0.367 & 0.445 \\ 
                \quad\quad\quad\quad\quad\rotatebox[origin=c]{180}{$\Lsh$}5-shot & 
                \quad\rotatebox[origin=c]{180}{$\Lsh$}0.662 &
                \quad\rotatebox[origin=c]{180}{$\Lsh$}0.378 \\ 
                
                Mistral-7B(0-shot) & 0.530 & 0.514 \\
                \quad\quad\quad\quad\quad\rotatebox[origin=c]{180}{$\Lsh$}5-shot & 
                \quad\rotatebox[origin=c]{180}{$\Lsh$}\textbf{0.693} & \quad\rotatebox[origin=c]{180}{$\Lsh$}\textbf{0.599} \\ 
                
                Qwen1.5-7B(0-shot) & 0.469 & 0.525 \\
                \quad\quad\quad\quad\quad\rotatebox[origin=c]{180}{$\Lsh$}5-shot &
                \quad\rotatebox[origin=c]{180}{$\Lsh$}0.606 & 
                \quad\rotatebox[origin=c]{180}{$\Lsh$}0.549 \\ 
        
                LLaMa3-8B(0-shot) & 0.636 & 0.549 \\ 
                \quad\quad\quad\quad\quad\rotatebox[origin=c]{180}{$\Lsh$}5-shot & 
                \quad\rotatebox[origin=c]{180}{$\Lsh$}0.544 & 
                \quad\rotatebox[origin=c]{180}{$\Lsh$}0.296 \\ 

                \midrule
                RoBERTa Fine-tune & 0.807 & 0.606 \\

                \bottomrule
            \end{tabular*}
        }
    \end{minipage}

    \label{table:utterance}
\end{table*}

\section{Findings}
\label{sec:results}

\subsection{Utterance-level Understanding}

Table~\ref{table:utterance} compares the performance of LLMs in child-adult utterance classification. There are 4 observations: 1) LLMs show low F1 scores in 0-shot settings for child-adult classification. The performance is higher in the ADOS dataset, possibly due to its more structured interaction design; 2) Increasing the conversation context to 5 (5-shot) can substantially improve classification, with Mistral-7B's performance on ADOS rising from 0.530 to 0.693; 3) However, adding more context does not consistently improve performance, as seen with the LLaMa3-8B's decrease with 5-shot across both datasets; 4) Lastly, we observe that best-performed LLMs (with 5-shot) yield close performance to fine-tuning baselines using RoBERTa.

\subsection{Session-level Understanding}

We present the performance of different LLMs in session-level understanding tasks in Table~\ref{table:transcript}.

\vspace{0.5mm}
\noindent \textbf{Activity/Observation Task.} Results show the highest F1 scores for activity/observation recognition are 0.961 and 0.482 in the ADOSMod3 and Remote-NLS datasets, respectively. The lower score in Remote-NLS is due to its more challenging multilabel task. Interestingly, increasing model size does not improve the performance, and LLaMa3-8B shows significant prediction variability.

\vspace{0.7mm}
\noindent \textbf{Language Skill.} We identify that the performance of language skill prediction is lower in the ADOSMod3 dataset for all models, with the highest F1 score being 0.579. One possible reason is that the ADOSMod3 transcriptions cover only two subtasks from the entire ADOS session. In contrast, when analyzing the Remote-NLS dataset, which includes full transcriptions, the best-performed LLM achieves an F1 score of 0.795. As the majority of children in remote-NLS datasets are with ASD, this suggests that LLMs can predict language skills for ASD, highlighting their promise to assist the clinician with language-related assessments.

\vspace{0.5mm}
\noindent \textbf{Individual Traits.} The highest F1 score for age predictions are 0.748 and 0.487 in ADOS-Mod3 and Remote-NLS datasets, respectively. Since the Remote-NLS dataset consists of children under 8 years old, these results highlight the challenges LLMs face in accurately predicting age ranges and potentially other traits for younger children.

\begin{table*}[t]
    \caption{Results of session-level understanding tasks including predicting activity, language skill, and demographics. \textbf{Bold} indicate the best results in each dataset. Activity recognition is a multilabel task in the Remote-NLS dataset.}
    \vspace{1mm}
    \footnotesize
    \begin{minipage}{0.57\textwidth}
        \centering
        \includegraphics[width=\linewidth]{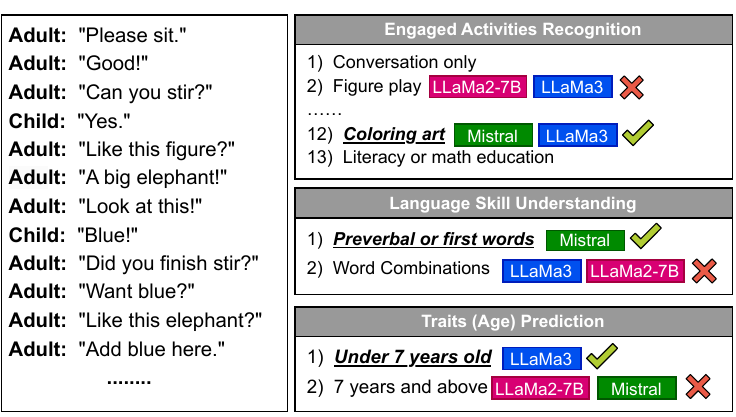}
    \end{minipage}
    \begin{minipage}{0.07\textwidth}
        \centering
    \end{minipage}
    \begin{minipage}[t]{0.43\textwidth}
        \centering
        \resizebox{\textwidth}{!}{
            \begin{tabular}{lccc}
    
                \toprule
                & \textbf{Activity} & \textbf{Language Skill} & \textbf{Traits} \\
                \midrule
                

                \textbf{ADOSMod3} & & & \\ 
                \quad LLaMa2-7B & 0.815 & 0.432 & 0.553 \\
                \quad Mistral-7B & 0.945 & \textbf{0.579} & 0.645 \\
                \quad Qwen1.5-7B & 0.895 & 0.542 & 0.592 \\
                \quad LLaMa3-8B & 0.353 & 0.452 & 0.625 \\
                \quad LLaMa2-13B & 0.818 & 0.452 & 0.378 \\
                \quad Qwen1.5-14B & \textbf{0.984} & 0.285 & \textbf{0.688} \\

                \midrule
                \textbf{Remote-NLS} & & & \\ 
                \quad LLaMa2-7B & 0.231 & 0.581 & 0.452 \\
                \quad Mistral-7B & 0.477 & \textbf{0.795} & 0.434 \\
                \quad Qwen1.5-7B & 0.408 & 0.261 & 0.415 \\
                \quad LLaMa3-8B & \textbf{0.482} & 0.226 & 0.415 \\
                \quad LLaMa2-13B & 0.270 & 0.439 & \textbf{0.487} \\
                \quad Qwen1.5-14B & 0.475 & 0.439 & 0.470 \\
                
                \bottomrule
            \end{tabular}
        }
    \end{minipage}
    \label{table:transcript}
\end{table*}

\begin{table}[t]
    \caption{Session-level understanding with majority voting of 3 best-performing LLMs (Metric: F1). $\uparrow$ indicates increased performance compared to a single LLM.}
    \small
    \centering
    \begin{tabular}{lccc}

        \toprule
        & \textbf{Activity} & \textbf{Language} & \textbf{Traits} \\
        \midrule
        
        \textbf{ADOSMod3} & 0.972 & 0.583 $\uparrow$ & 0.722 $\uparrow$ \\ 
        \textbf{Remote-NLS} & 0.527 $\uparrow$ & 0.817 $\uparrow$ & 0.500 $\uparrow$ \\

        \bottomrule
    \end{tabular}
    \label{table:combine}
\end{table}

\begin{table}[t]
    \caption{Results of session-level understanding tasks including predicting activity, language skill, and demographics with chain-of-thought (COT) prompt.}
    \footnotesize
    \centering
    \begin{tabular}{lccc}
    
        \toprule
        & \textbf{Activity} & \textbf{Language Skill} & \textbf{Traits} \\
        \midrule
        
        \textbf{ADOSMod3} & & & \\ 
        \quad LLaMa2-7B & 0.420 & 0.434 & 0.424 \\
        \quad Mistral-7B & 0.875 & 0.538 & 0.597 \\
        
        \midrule
        \textbf{Remote-NLS} & & & \\ 
        \quad LLaMa2-7B & 0.277 & 0.517 & 0.457 \\
        \quad Mistral-7B & 0.395 & 0.361 & 0.434 \\
        \bottomrule
    \end{tabular}
    \label{table:cot}
\end{table}

\begin{figure}[t]
  \centering
  \includegraphics[width=0.75\linewidth]{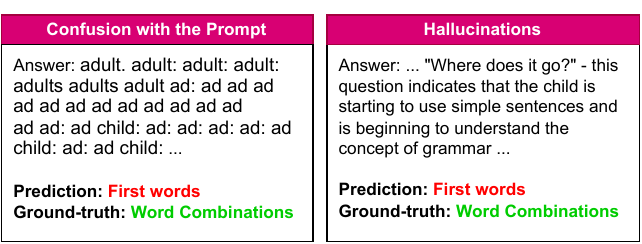}
  \caption{Exemplar output including confusion and hallucinations from the LLaMa2-7B in predicting language skills.}
  \label{fig:error}
\end{figure}

\subsection{How To Improve Predictions of LLMs?}
We continue investigating whether more advanced prompts, e.g., chain-of-thought (COT) \cite{wei2022chain}, or majority voting with top-performing LLMs improve performance. The majority voting results are in Table~\ref{table:combine}, and the COT prompt results are in Table~\ref{table:cot}. From Table~\ref{table:combine}, we can observe that combining output from different LLMs effectively improves the prediction performance. However, from the comparisons of the COT results, we observe that adding COT to the prompt consistently lowers the performance of our application compared to the 0-shot prompt without COT. Human inspections of the model output indicate that while COT prompts aim to elicit reasons for their choices, they also lead to increases in the hallucinations in the model output.


\subsection{Concerns with LLMs}
Apart from the promises that LLMs bring for automated analysis of child-adult interactions, we identify two primary sources of errors in the model's output in different prediction tasks. The first one is that the model may sometimes have confusion with the prompt, leading to the generation of random words, as presented in Fig~\ref{fig:error}. Moreover, as shown in Figure~\ref{fig:error}, LLMs output incorrect responses as they tend to hallucinate with certain phrases \cite{zhang2023siren} when making predictions to language skills.

\begin{table}[t]
    \caption{Results between non-experts (average performance) and best-performed LLMs on randomly selected test samples from the Remote-NLS dataset (Metric: F1).}
    \footnotesize
    \centering
    \begin{tabular}{p{1.7cm}cccc}

        \toprule
        & \textbf{Utterance} & \textbf{Activity} & \textbf{Language} \\
        \midrule
        
        \textbf{Non-experts} & \textbf{0.717} & 0.332 & 0.592 \\ 
        \textbf{Best LLM} & 0.650 & \textbf{0.448} & \textbf{1.000} \\

        \bottomrule
    \end{tabular}
    \label{table:human}
\end{table}

\subsection{Can LLMs Outperform Non-experts?}
Table~\ref{table:human} compares the best-performing LLM with human evaluations on randomly selected samples. We observe that LLMs outperform non-expert humans on both activity (Human: 0.332; LLM: 0.448) and language skills (Human: 0.592; LLM: 1.000) predictions from the test data. Though humans outperform LLMs in classifying utterances, the difference is minimal. Moreover, evaluation feedback in Section~\ref{app:feedback} highlights the notable efforts required to perform the task. These results show the ability of LLMs to help segment interactions of interest and evaluate language skills in our context.

\subsection{Feedback from Non-expert Evaluations}
\label{app:feedback}
Non-experts feedback on manually performing the task of analyzing transcriptions (both utterance level and session level) of adult-child interactions reveals several significant challenges. 

\noindent 1) First, the process is notably labor-intensive, requiring sustained cognitive engagement and attention to detail, which can lead to fatigue and the potential to make more erroneous decisions; 

\noindent 2) Second, the complexity of tracking conversational sequences, particularly when adults self-respond to questions, complicates the accurate attribution of utterances either to child or adult (i.e., the task of child-adult utterance classification); 

\noindent 3) Third, non-verbal responses in interactions introduce additional interpretative challenges, as children might respond with gestures rather than verbal replies, requiring inference beyond the transcribed content (specifically for activity/observation task). This necessitates a comprehensive understanding of the interaction context, which is often difficult to derive from transcriptions alone;

\noindent 4) Finally, assessing children's language development (i.e., the task of predicting language skills) is further complicated by variability in verbal expression, with children often oscillating between mostly single words and sometimes word combinations. Age-related discrepancies in language skills add to the complexity, as some younger children may exhibit better verbal abilities compared to some older children due to different conditions related to ASD. These issues highlight the demanding nature of manual analysis and the potential for variability and inaccuracies in human assessments.

\section{Conclusion}

In this work, we evaluate the ability of generic LLMs to analyze interactions between children with autism and adults in clinical observations. Our findings reveal that: 1) Current LLMs show promise in understanding child-adult clinical interactions in both utterance-level and session-level predictions, implying their potential to assist clinicians; 2) LLMs can perform comparably or even outperform non-expert humans in designed tasks, including activity and language recognition; 3) However, we observe hallucinations and confusions in LLMs, which raises concerns for their wide deployment.


Moreover, our current evaluation only considers transcripts from child-adult dyadic interactions in clinical assessment, while other modalities, such as speech and videos, are not included. In the future, we also plan to use speech foundation models and vision language models (VLMs) to provide multi-modal evaluations along with LLMs. On the other hand, as LLMs are known to have biased output, one limitation of our work is the lack of analysis of biases of LLMs in performing these tasks. Lastly, although we observe hallucinations in LLM's response, we have not performed a systematic study on how these hallucinated outputs would impact the assessment from experts. Such a study could shed light on the design of future LLMs to mitigate hallucinations and improve their reliability in clinical and diagnostic settings involving children with autism.

\section{Limitations and Future Directions}
While our work provides valuable baselines and insights into four tasks related to understanding child-adult dyadic interactions, it lacks a more in-depth analysis of the prediction of ASD. One major reason is the need for more high-quality transcriptions for the experiments, such as in the ADOSMod3 dataset. One future direction is to enable the state-of-the-art ASR \cite{fan24b_interspeech} and speaker diarization \cite{xu24c_interspeech, xu2024data} using speech foundation models to acquire more transcriptions with speaker labels for our evaluation. Another limitation of our work is that our experiments rely only on expert transcriptions, excluding those from ASR services. One future extension would compare the current results with those using ASR transcriptions.

Moreover, our current evaluation only considers transcripts from child-adult dyadic interactions in clinical assessment, while other modalities, such as speech and videos, are not included. In the future, we also plan to use speech foundation models and vision language models (VLMs) to provide multi-modal evaluations along with LLMs. On the other hand, as LLMs are known to have biased output, one limitation of our work is the lack of analysis of biases of LLMs in performing these tasks \cite{liang2021towards}. Lastly, although we observe hallucinations in LLM's response, we have not performed a systematic study on how these hallucinated outputs would impact the assessment from experts and approaches to reduce the hallucinations \cite{liu2024reducing}. Such a study could shed light on the design of future LLMs to mitigate hallucinations and improve their reliability in clinical and diagnostic settings involving children with autism.

\bibliographystyle{plain}
\bibliography{ref}

\begin{thebibliography}{10}

\bibitem{bai2023qwen}
Jinze Bai, Shuai Bai, Yunfei Chu, Zeyu Cui, Kai Dang, Xiaodong Deng, Yang Fan, Wenbin Ge, Yu~Han, Fei Huang, et~al.
\newblock Qwen technical report.
\newblock {\em arXiv preprint arXiv:2309.16609}, 2023.

\bibitem{barokova2020commentary}
Mihaela Barokova et~al.
\newblock Commentary: Measuring language change through natural language samples.
\newblock {\em Journal of autism and developmental disorders}, 50(7):2287--2306, 2020.

\bibitem{bishop2017autism}
Somer~L Bishop, Marisela Huerta, Katherine Gotham, Karoline Alexandra~Havdahl, Andrew Pickles, Amie Duncan, Vanessa Hus~Bal, Lisa Croen, and Catherine Lord.
\newblock The autism symptom interview, school-age: A brief telephone interview to identify autism spectrum disorders in 5-to-12-year-old children.
\newblock {\em Autism Research}, 10(1):78--88, 2017.

\bibitem{bone2015applying}
Daniel Bone, Matthew~S Goodwin, Matthew~P Black, Chi-Chun Lee, Kartik Audhkhasi, and Shrikanth Narayanan.
\newblock Applying machine learning to facilitate autism diagnostics: pitfalls and promises.
\newblock {\em Journal of autism and developmental disorders}, 45:1121--1136, 2015.

\bibitem{butler2022remote}
Lindsay~K Butler, Chelsea La~Valle, Sophie Schwartz, Joseph~B Palana, Cerelia Liu, Natalie Peterman, Lue Shen, and Helen Tager-Flusberg.
\newblock Remote natural language sampling of parents and children with autism spectrum disorder: Role of activity and language level.
\newblock {\em Frontiers in Communication}, 7:820564, 2022.

\bibitem{fan24b_interspeech}
Ruchao Fan, Natarajan {Balaji Shankar}, and Abeer Alwan.
\newblock Benchmarking children's asr with supervised and self-supervised speech foundation models.
\newblock In {\em Interspeech 2024}, pages 5173--5177, 2024.

\bibitem{hessel2023androids}
Jack Hessel, Ana Marasovi{\'c}, Jena~D Hwang, Lillian Lee, Jeff Da, Rowan Zellers, Robert Mankoff, and Yejin Choi.
\newblock Do androids laugh at electric sheep? humor “understanding” benchmarks from the new yorker caption contest.
\newblock In {\em Proceedings of the 61st Annual Meeting of the Association for Computational Linguistics (Volume 1: Long Papers)}, pages 688--714, 2023.

\bibitem{jiang2023mistral}
Albert~Q Jiang, Alexandre Sablayrolles, Arthur Mensch, Chris Bamford, Devendra~Singh Chaplot, Diego de~las Casas, Florian Bressand, Gianna Lengyel, Guillaume Lample, Lucile Saulnier, et~al.
\newblock Mistral 7b.
\newblock {\em arXiv preprint arXiv:2310.06825}, 2023.

\bibitem{liang2021towards}
Paul~Pu Liang, Chiyu Wu, Louis-Philippe Morency, and Ruslan Salakhutdinov.
\newblock Towards understanding and mitigating social biases in language models.
\newblock In {\em International Conference on Machine Learning}, pages 6565--6576. PMLR, 2021.

\bibitem{liu2024reducing}
Sheng Liu, Haotian Ye, and James Zou.
\newblock Reducing hallucinations in vision-language models via latent space steering.
\newblock {\em arXiv preprint arXiv:2410.15778}, 2024.

\bibitem{liu2019roberta}
Yinhan Liu, Myle Ott, Naman Goyal, Jingfei Du, Mandar Joshi, Danqi Chen, Omer Levy, Mike Lewis, Luke Zettlemoyer, and Veselin Stoyanov.
\newblock Roberta: A robustly optimized bert pretraining approach.
\newblock {\em arXiv preprint arXiv:1907.11692}, 2019.

\bibitem{lord2020autism}
Catherine Lord, Traolach~S Brugha, Tony Charman, James Cusack, Guillaume Dumas, Thomas Frazier, Emily~JH Jones, Rebecca~M Jones, Andrew Pickles, Matthew~W State, et~al.
\newblock Autism spectrum disorder.
\newblock {\em Nature reviews Disease primers}, 6(1):1--23, 2020.

\bibitem{lord2000autism}
Catherine Lord, Susan Risi, Linda Lambrecht, Edwin~H Cook, Bennett~L Leventhal, Pamela~C DiLavore, Andrew Pickles, and Michael Rutter.
\newblock The autism diagnostic observation schedule—generic: A standard measure of social and communication deficits associated with the spectrum of autism.
\newblock {\em Journal of autism and developmental disorders}, 30:205--223, 2000.

\bibitem{tager2009defining}
Helen Tager-Flusberg, Sally Rogers, Judith Cooper, Rebecca Landa, Catherine Lord, Rhea Paul, Mabel Rice, Carol Stoel-Gammon, Amy Wetherby, and Paul Yoder.
\newblock Defining spoken language benchmarks and selecting measures of expressive language development for young children with autism spectrum disorders.
\newblock 2009.

\bibitem{touvron2023llama}
Hugo Touvron, Louis Martin, Kevin Stone, Peter Albert, Amjad Almahairi, Yasmine Babaei, Nikolay Bashlykov, Soumya Batra, Prajjwal Bhargava, Shruti Bhosale, et~al.
\newblock Llama 2: Open foundation and fine-tuned chat models.
\newblock {\em arXiv preprint arXiv:2307.09288}, 2023.

\bibitem{volden1991neologisms}
Joanne Volden et~al.
\newblock Neologisms and idiosyncratic language in autistic speakers.
\newblock {\em Journal of autism and developmental disorders}, 21(2):109--130, 1991.

\bibitem{wang2024diarizationlm}
Quan Wang, Yiling Huang, Guanlong Zhao, Evan Clark, Wei Xia, and Hank Liao.
\newblock Diarizationlm: Speaker diarization post-processing with large language models.
\newblock {\em arXiv preprint arXiv:2401.03506}, 2024.

\bibitem{wei2022chain}
Jason Wei, Xuezhi Wang, Dale Schuurmans, Maarten Bosma, Fei Xia, Ed~Chi, Quoc~V Le, Denny Zhou, et~al.
\newblock Chain-of-thought prompting elicits reasoning in large language models.
\newblock {\em Advances in neural information processing systems}, 35:24824--24837, 2022.

\bibitem{xu2024data}
Anfeng Xu, Tiantian Feng, Helen Tager-Flusberg, Catherine Lord, and Shrikanth Narayanan.
\newblock Data efficient child-adult speaker diarization with simulated conversations.
\newblock {\em arXiv preprint arXiv:2409.08881}, 2024.

\bibitem{xu24c_interspeech}
Anfeng Xu, Kevin Huang, Tiantian Feng, Lue Shen, Helen Tager-Flusberg, and Shrikanth Narayanan.
\newblock Exploring speech foundation models for speaker diarization in child-adult dyadic interactions.
\newblock In {\em Interspeech 2024}, pages 5193--5197, 2024.

\bibitem{zhang2023siren}
Yue Zhang, Yafu Li, Leyang Cui, Deng Cai, Lemao Liu, Tingchen Fu, Xinting Huang, Enbo Zhao, Yu~Zhang, Yulong Chen, et~al.
\newblock Siren's song in the ai ocean: a survey on hallucination in large language models.
\newblock {\em arXiv preprint arXiv:2309.01219}, 2023.

\bibitem{zhao2023survey}
Wayne~Xin Zhao, Kun Zhou, Junyi Li, Tianyi Tang, Xiaolei Wang, Yupeng Hou, Yingqian Min, Beichen Zhang, Junjie Zhang, Zican Dong, et~al.
\newblock A survey of large language models.
\newblock {\em arXiv preprint arXiv:2303.18223}, 2023.

\end{thebibliography}


\end{document}